\documentclass{article}

\usepackage{times}
\usepackage{graphicx} % more modern
\usepackage{subfigure} 

\usepackage{natbib}

\usepackage{algorithm}
\usepackage{algorithmic}

\usepackage{hyperref}

\usepackage[accepted]{icml2016}

\icmltitlerunning{Lazy Evaluation of Convolutional Filters}

\begin{document} 

\twocolumn[
\icmltitle{Lazy Evaluation of Convolutional Filters}

\icmlauthor{Sam Leroux, Steven Bohez, Cedric De Boom, Elias De Coninck,\\ Tim Verbelen, Bert Vankeirsbilck, Pieter Simoens, Bart Dhoedt}{first.lastname@ugent.be}
\icmladdress{Ghent University - iMinds, Department of Information Technology,  
            Technologiepark-Zwijnaarde 15, 9052 Gent, Belgium}

\icmlkeywords{}

\vskip 0.3in
]

\begin{abstract} 
In this paper we propose a technique which avoids the evaluation of certain convolutional filters in a deep neural network. This allows to trade-off the accuracy of a deep neural network with the computational and memory requirements. This is especially important on a constrained device unable to hold all the weights of the network in memory.
\end{abstract} 

\section{Introduction}
\label{introduction}
Deep neural networks are good candidates to enable the next generation of pervasive devices. Internet-of-Things (IoT) devices are commonplace in our everyday lives, yet  are still limited in their functionality. Combining the intelligence of deep neural networks with vast amounts of rich sensor data available in an IoT ecosystem could allow for a truly Internet-of-Smart-Things.
\\
\newline
Deep neural networks require large amounts of resources, both to train and to evaluate. Training is usually less of a problem since this can be done offline on large GPU clusters in the cloud. Inference on the other hand is more of a challenge. The typical IoT devices are limited in the resources available, they usually contain a low-power single-core CPU, limited memory and are often battery powered. Evaluating the current state-of-the-art deep neural networks on these devices is often simply not possible.
\\
\newline
Current state-of-the-art architectures are usually deep and wide. Impressive results have been obtained by converting these large trained networks into smaller, computationally less expensive versions. \cite{ba2014deep, romero2014fitnets}.
\newpage
It is well known that 32 bit floating point numbers are not needed, 16 bit \cite{gupta2015deep}, 10 bit \cite{courbariaux2014training}, 8 bit \cite{vanhoucke2011improving} and even binary \cite{courbariaux2016binarynet} and fixed point precision \cite{lin2015fixed} weights and activations are sufficient for training and evaluating a neural network.
\\
\newline
Another approach is presented in \cite{chen2015compressing} where the authors use a hash function to group connection weights into hash buckets. All connections with the same hash value share the same parameter value thereby reducing the number of parameters to store. Other techniques to exploit the redundancy among weights include low-rank decompositions of the weight matrices \cite{sainath2013low, denil2013predicting, sindhwani2015structured} and sparsity inducing regularisation techniques \cite{collins2014memory}.
\\
\newline
Other approaches optimize the structure of the neural network itself. A three step method is presented in \cite{han2015learning} where first the network is trained to discover which connections are important, then, the redundant connections are pruned and finally the network is retrained to fine-tune the weights of the remaining connections. This procedure is able to reduce the number of parameters up to 13 times without any loss of accuracy.
\\
\newline
In this paper we present a lazy evaluation approach which allows reducing the required runtime of a deep neural network by selectively evaluating the convolutional filters. Our approach is most similar to the perforatedCNNs technique \cite{figurnov2015perforatedcnns} which avoids evaluating convolutional filters for some of the spatial positions. The filters are only evaluated for  a subset of the spatial positions, an interpolated value is used for the other positions. Our approach on the other hand evaluates the filters at every spatial position but reduces the number of filters that need to be evaluated. A combination of both techniques could allow for an even larger reduction in computational cost since both techniques exploit orthogonal properties of the network.
\\
\newline
The remainder of this paper is organised as follows: In Section \ref{concept} we introduce the concept of Lazy evaluation of convolutional filters. In Section \ref{experiments} we present the experimental results. We conclude in Section \ref{conclusion and future work} with the future work.

\section{Concept}
\label{concept}
One interesting property of deep convolutional neural networks (CNNs) is that they learn a hierarchy of features \cite{le2013building,razavian2014cnn, yosinski2014transferable}. The first layers learn to detect low level features such as oriented edges and color transitions. These features are then combined by the deeper layers into high level concepts such as human faces and various objects. 
\\
\newline
The default implementation of a CNN evaluates every filter of every layer as the input is being processed by the network. Filters that are not relevant will return a feature map with extremely small values and will have little impact on the final classification. We try to prune these irrelevant filters on a per sample basis by predicting which filters will be useful for the specific input based on the activations of the filters in the previous convolutional layer at runtime.
\\
\newline
We define the \textbf{Activation Strength} of a certain convolutional filter when processing the input as the sum of the absolute values in the output of the filter. We hypothesise that only filters with a large Activation Strength have an impact on the final classification. Consequently, the filters with the lowest Activation Strength are not relevant at all and can be omitted, effectively setting the activation for the entire feature map to zero instead of performing the actual computation.
\\
\newline
We use linear regression to predict the Activation Strength of the filters in layer $i$ based on the Activations of the filters in layer $i-1$.

The forward propagation algorithm is changed as shown in algorithm \ref{alg:fprop}
\begin{algorithm}[h]
   \caption{Forward propagation through the network}
   \label{alg:fprop}
\begin{algorithmic}
   \FOR{each layer $l$ in the network}
   \IF{$l$ is a convolutional layer} 
   \STATE $\bullet$ Use linear regression to predict the activation strengths of layer $l$ based on the activation strengths of layer $l-1$
   \STATE $\bullet$ Evaluate the $n_l\%$ filters with the largest predicted Activation Strengths, use zero values for the other filters
   \ELSE
   \STATE Use the unmodified forward propagation for this layer
   \ENDIF
   \ENDFOR
\end{algorithmic}
\end{algorithm}
The additional hyperparameter $n_l$ for each convolutional layer $l$ is used to trade-off accuracy and computational cost at runtime. The ability to dynamically trade off accuracy and computation is especially interesting for mobile devices that are battery operated. A suitable trade-off parameter can be selected based on the remaining battery capacity and the remaining operation time or on the desired runtime and accuracy.

\section{Experiments}
\label{experiments}
In this section we present the preliminary results of our approach. We choose an image classification task since this is arguably the current benchmark for deep convolutional neural networks and because of the high dimensional input data which requires large amounts of memory and computation power. We focus on real-time image processing and propagate the dataset one image at a time through the network in all these experiments. This best resembles the real world applications where data has to be processed the moment it becomes available. There is no time to accumulate images in batch to allow for optimized batch processing.
\\
\newline
We choose the VGG network \cite{simonyan2014very} with 19 layers trained on the Imagenet \cite{deng2009imagenet} dataset as the base network to optimize since this is a typical, widely used architecture obtaining near state-of-the-art performance. All experiments were implemented in Theano \cite{bergstra+al:2010-scipy}. All timings reported are measured on an Intel i5-2400 CPU.
%The full VGG network requires over 500MB of storage simply to store the weights. 
\\
\newline
The following sections are organised following the different research questions posed in this research.
\subsection{What is the impact of pruning convolutional filters on the accuracy and runtime of the network ?}
\subsubsection{Impact on accuracy}
\begin{figure}[t]
\begin{center}
\centerline{
\includegraphics[width=\columnwidth, height=235px]{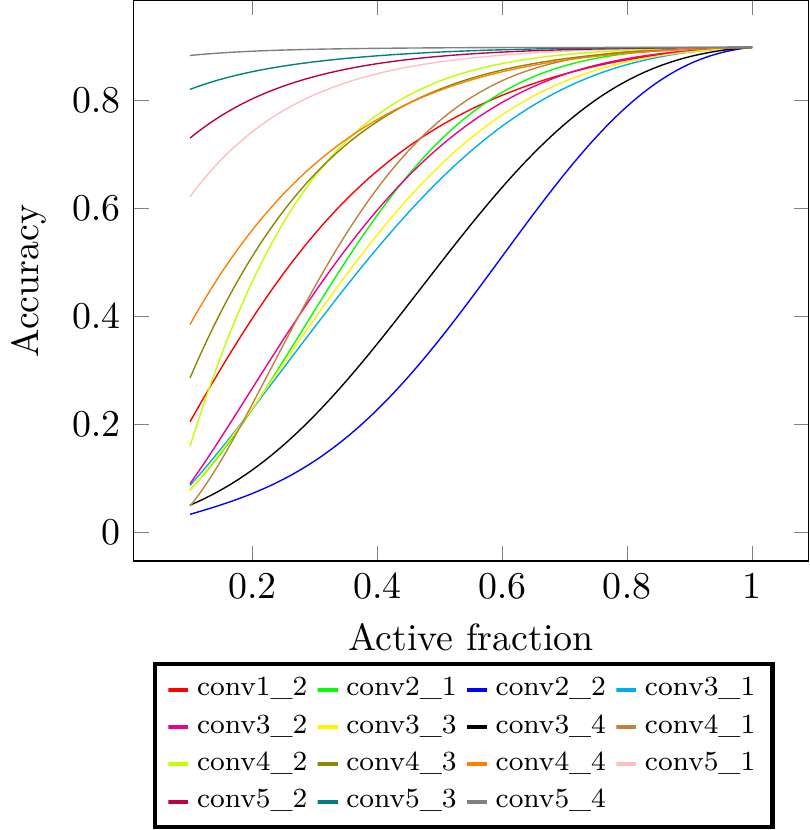}
}
\caption{The global accuracy of the network as a function of the active filters for each layer. This graph illustrates how sensitive the accuracy is to ignoring filters in each convolutional layer. The layer names follow the original VGG paper \cite{simonyan2014very}}
\label{fig:accuracyperfilter}
\end{center}
\end{figure}

We processed each image in our validation set and for each image and each convolutional layer we independently set the $n_l\%$ activations with the smallest activation strength to zero and recorded the accuracy of the entire network. We expect that the deep layers have highly specialised filters and that the majority of these filters can be ignored while still allowing a high classification accuracy. The filters in the first layers, on the other hand, are low level filters and should each have a useful contribution to the accuracy of the network.
\\
\newline
The results are shown in Figure \ref{fig:accuracyperfilter}. We observe the predicted behaviour although less straight forward than expected. The last layer in the network for example is highly specialised, up to 80\% of its filters (total of 512 filters) can be ignored without any significant drop in accuracy ($-0.4\%$). Dropping filters from one of the first convolutional layers incurs a much higher penalty. Not all layers follow this global trend. The ``conv2\_2'' layer for example is the most sensitive to ignored filters, even more sensitive than the very first convolutional layer. 

\subsubsection{Impact on runtime}
\begin{figure}[t]
\begin{center}
\includegraphics[width=\columnwidth,height=235px]{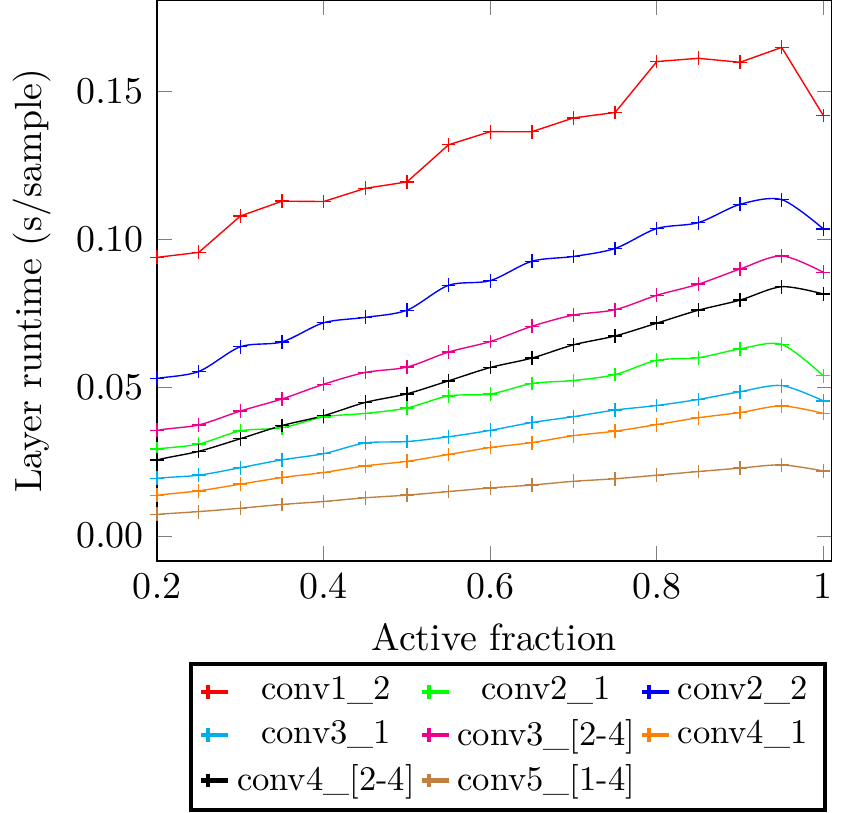}
\caption{The required runtime of each convolutional layer as a function of the active filters, measured on a single core CPU.}
\label{fig:runtime}
\end{center}
\end{figure}

The previous section showed that we can ignore certain filters in each convolutional layer without a significant drop in accuracy. In this section we investigate the impact on the required runtime. The results are shown in Figure \ref{fig:runtime}. This graph shows the computational cost for each convolutional layer as a function of the fraction of active filters. We observe a more or less linear relationship. The sudden drop in computational cost when all filters are used ($x=1$) is caused by a suboptimal implementation where data needs to be copied into a preallocated buffer. This is especially costly for the early layers since these produce the largest activation maps. A more efficient in-place implementation should solve this. The overhead of predicting the most important filters based on the activations of the previous layer is included in these measurements and is small compared to the cost of evaluating all convolutional filters ($<2\%$).
\\
\newline
We only show the results measured on a single CPU core. Graphical Processing Units (GPUs) allow for a much more efficient evaluation of a neural network because of their inherent parallelism \cite{krizhevsky2012imagenet}. The approach presented in this paper is not useful for GPU implementations because the cost of evaluating extra filters on a GPU is relatively small compared to the cost caused by transfering data to and from the device. In an IoT use case however most of the devices are single core CPU operated and could benefit from the lazy filter evaluation approach. This is especially true when the network is too large to fit into the memory of the device and the filters need to be processed sequentially. Secondary memory access is needed in those cases to retrieve the weights while processing data.
\\
\newline
The last convolutional layer (``conv5\_4'') is the least sensitive to ignored filters and as such a good candidate for heavy pruning. The computational cost of this layer is however only a small part of the total computational cost ($\approx 3\%$). It is still useful to prune most of the ``conv5\_4'' filters since this results in a very sparse activation map which allows a large speed-up of the fully connected layer following this layer and an even larger reduction in required memory (see Section \ref{section_memory}). 

\subsection{Can we predict the relevant filters of a certain layer based on the activations of the previous \ \ layer ?}
The crucial part of this technique is predicting which filters will be relevant before evaluating them. We used linear regression to predict the Activation Strength of each filter in a convolutional layer based on the Activation Strengths of the filters in the previous convolutional layer.
\\\[\mathbf{s_i} = \mathbf{s_{i-1}} \cdot \mathbf{W} + \mathbf{b}\] where $\mathbf{s_i}$ is a vector of dimensionality $m$ (the number of convolutional filters in layer $i$), $\mathbf{s_{i-1}}$ is a vector of dimensionality $n$ (the number of filters in the previous convolutional layer, $\mathbf{W}$ is an $m*n$ weight matrix and $\mathbf{b}$ is an $m$-dimensional bias vector. We used gradient descent to minimise the mean absolute error between the predicted and the real activation strengths. On average we are able to correctly predict about 90\% of the top N\% filters for each layer.

%\begin{figure}[h]
%\begin{center}
%\includegraphics[width=\columnwidth]{img/predicting.png}
%\caption{The required runtime of the network as a function of the active filters, measured on the CPU.}
%\label{fig:predicting}
%\end{center}
%\end{figure}

\subsection{What is the gain in runtime and the loss in accuracy of this approach ?}
\begin{figure}[h]
\begin{center}
\includegraphics[width=\columnwidth]{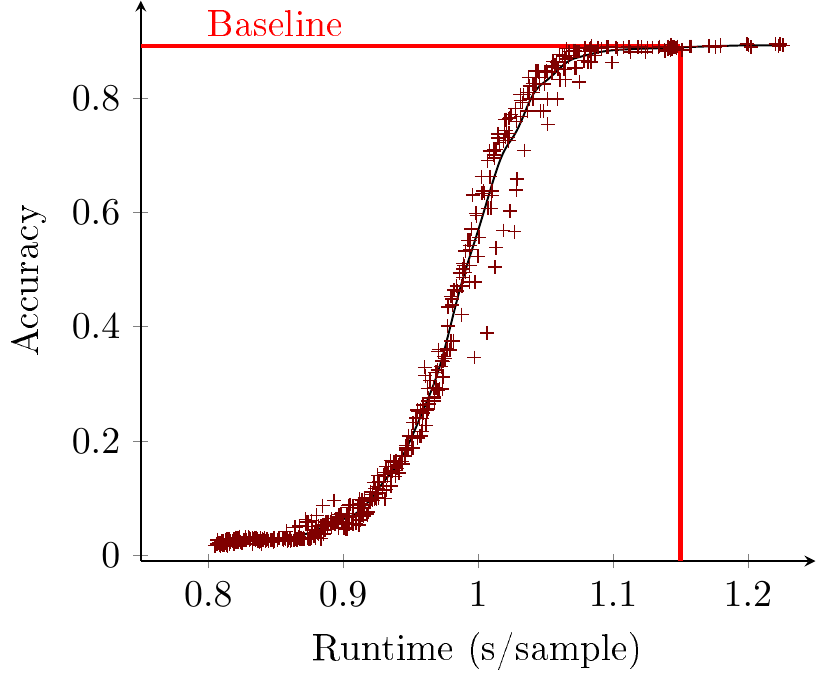}
\caption{The trade-off between accuracy and required runtime.}
\label{fig:tradeoff}
\end{center}
\end{figure}

The technique presented in this paper allows a runtime trade-off between accuracy and speed. The task of finding suitable trade-off parameters is a multi objective optimization task characterized by a Pareto front. We used the NSGA-II algorithm \cite{deb2002fast} implemented in PyGMO \cite{izzo2012pygmo} to explore this Pareto front. The result is presented in Figure \ref{fig:tradeoff}. The horizontal and vertical lines show the baseline accuracy, respectively the baseline runtime of the network.

\subsection{How can we use this technique to reduce the memory footprint of the network?}
\label{section_memory}
The computational cost of a convolutional neural network is dominated by the convolutional layers. The fully connected layers on the other hand dictate the memory footprint. The first fully connected layer in the VGG19 network for example has 102764544 parameters and needs 411MB just to store these weights (float32). Figure \ref{fig:accuracyperfilter} shows that the last convolutional layer (``conv5\_4'') is highly specialised, up to 80\% of the filters can be ignored without any significant impact on accuracy. This results in a very sparse activation map. Figure \ref{fig:runtime} showed that disabling these filters unfortunately has little impact on the required runtime of this layer. The memory footprint of the first fully connected layer however is directly proportional to the number of active filters in the last convolutional layer. When we disable 80\% of the filters only 20\% of the weights of the fully connected layer are needed since the other 80\% will be multiplied with zero values. We only need to load a subset of the weights into memory at runtime (i.e. 88MB instead of 441MB) thanks to the sparsity of the last convolutional layer.

\section{Conclusion and future work}
\label{conclusion and future work}
We presented an approach which avoids evaluating convolutional filters that are unlikely to have an impact on the final classification. We trained a linear regression model for each convolutional layer to predict the importance of each convolutional filter based on the activations of the previous layer. This allowed us to prune low-impact filters at runtime. on a per-sample basis. As a consequence the activations can be very sparse reducing the number of parameters that need to be retrieved from secondary storage mediums on devices that are unable to hold all parameters in memory.
\\
\newline
In future work we will investigate if it is possible to combine this approach with the techniques presented in the related work section. We will also implement this technique on embedded and FPGA platforms where the weights do not fit in on-chip memory and external memory access is the bottleneck  during computation.

\section*{Acknowledgment}
Part of this work was supported by the iMinds IoT Research Program. Steven Bohez is funded by a PhD grant of the Agency for Innovation by Science
and Technology in Flanders (IWT). Cedric De Boom is funded by a PhD grant of the Flanders Research Foundation (FWO). We gratefully acknowledge the support of NVIDIA Corporation with the donation of a Tesla K40 GPU and  a Jetson TK1 and TX1.

\clearpage
\bibliography{sparsity}
\bibliographystyle{icml2016}

\end{document}